\documentclass[twoside,leqno,twocolumn]{article}

\usepackage[letterpaper]{geometry}

\usepackage{siamproceedings}

\usepackage[T1]{fontenc}
\usepackage{amsfonts}
\usepackage{graphicx}
\usepackage{epstopdf}
\usepackage{enumitem}
\usepackage{algorithmic}
\ifpdf
  \DeclareGraphicsExtensions{.eps,.pdf,.png,.jpg}
\else
  \DeclareGraphicsExtensions{.eps}
\fi

\newsiamremark{remark}{Remark}
\newsiamremark{hypothesis}{Hypothesis}
\crefname{hypothesis}{Hypothesis}{Hypotheses}
\newsiamthm{claim}{Claim}

\usepackage{amsopn}

\usepackage{multirow}
\usepackage{booktabs}
\usepackage{xcolor}

\begin{document}

\title{\Large PIER: Physics-Informed Environmental Retrieval for Time-Series Modeling}
\author{
\parbox{\textwidth}{
\centering
Shiyuan Luo$^{1}$,
Runlong Yu$^{2}$,
Chonghao Qiu$^{1}$,
Yue Qin$^{1}$,
Rahul Ghosh$^{3}$,\\
Robert Ladwig$^{4}$,
Paul C. Hanson$^{5}$,
Yiqun Xie$^{6}$,
Xiaowei Jia$^{1}$\\[0.5em]
{\small
$^{1}$University of Pittsburgh \quad
$^{2}$University of Alabama \quad
$^{3}$University of Minnesota}\\
{\small
$^{4}$Aarhus University \quad
$^{5}$University of Wisconsin--Madison \quad
$^{6}$University of Maryland}\\[0.3em]
{\small
$^{1}$\texttt{\{shl298,chq29,yuq9,xiaowei\}@pitt.edu} \quad
$^{2}$\texttt{ryu5@ua.edu}}\\
{\small
$^{3}$\texttt{ghosh128@umn.edu} \quad
$^{4}$\texttt{rladwig@ecos.au.dk} \quad
$^{5}$\texttt{pchanson@wisc.edu} \quad
$^{6}$\texttt{xie@umd.edu}}\\[0.3em]
}
}
\date{}

\maketitle







\begin{abstract}
Accurate modeling of environmental systems is fundamental to scientific understanding and decision-making, yet remains challenging because observations are limited and physical dynamics vary across systems.
Retrieval-augmented approaches offer a natural path to transfer knowledge across systems, but standard embedding-based retrieval does not guarantee consistency of underlying physical processes, since scenarios with similar embeddings may arise from different underlying mechanisms. We propose Physics-Informed Environmental Retrieval (PIER), a model-agnostic framework that augments embedding-based retrieval with a physics-aware stream that scores candidates by flux-response consistency with the target, using local verifiers trained on physics-derived flux features. A weight adjustment mechanism then learns per-scenario weights that adaptively balance the two retrieval streams based on diagnostic features summarizing physics-stream reliability. Experiments on 356 lakes across the Midwestern United States spanning 41 years show that PIER consistently outperforms baselines for water temperature and dissolved oxygen prediction, and serves as a general augmentation strategy across diverse backbones.
\end{abstract}

\section{Introduction.}
Accurate modeling of environmental systems is fundamental to advancing scientific understanding of the natural world. Environmental processes span a wide range of spatial and temporal scales and involve nonlinear interactions among physical, chemical, and biological components, making direct observation and experimentation alone insufficient for comprehensive analysis and understanding. 
Computational models bridge this gap by integrating observational data with theoretical knowledge to simulate system behavior under varying scenarios and support decision-making.
In particular, the scientific community has long relied on process-based physical models that encode fundamental physical principles, such as conservation of mass, energy, and momentum, to simulate system behavior. Such models are widely used across aquatic science~\cite{hipsey2019general,ladwig2022long}, agriculture~\cite{jia2019bringing}, and geology~\cite{reichstein2019deep}. A key advantage of these models is that they produce physically meaningful intermediate states and flux quantities that reflect the governing processes of each system, providing interpretable and mechanistically grounded representations of system
dynamics. However, their predictive accuracy is often limited by imperfect and simplified parameterization, which can introduce systematic biases and constrain model fidelity~\cite{jia2019physics}. 

Machine learning (ML) offers a complementary path. By learning directly from data, ML models can capture nonlinear relationships and complex temporal patterns, and have shown strong results across a range of environmental applications~\cite{zhong2021machine, yu2025physics, wang2025carbonglobe, luo2026great}.
However, in real-world environmental problems, observational data are often limited, since consistent environmental monitoring requires expensive field studies, infrastructure, and ongoing maintenance, making comprehensive observation networks economically infeasible~\cite{willard2022integrating, zhuang2020comprehensive}. 
A common strategy is to train a single global model using data from a large number of observed sites and over long time periods. 
Such a global model works well in general, but real-world environmental problems exhibit complex and nonstationary patterns whose periodicities and dynamics vary across spatiotemporal contexts~\cite{han2025retrieval, xie2021statistically, kratzert2019towards}. 
For example, the temporal dynamics of water quality vary substantially across lakes owing to differences in geometric characteristics (e.g., depth, and area), trophic state, and local meteorological forcing; likewise, hydrological responses in watersheds are shaped by heterogeneous geological substrates, land cover distributions, and climate regimes~\cite{zhou2023controls, zhang2017global}.
More importantly, some of these conditions may not be observable and thus cannot be included in data. 
While a global model benefits from broad training coverage, it tends to converge toward an averaged representation that can obscure the localized dynamics unique to each individual system. On the other hand, separate modeling of each system is often constrained by sparse observation data. 
These limitations highlight the need for models that capture both the shared structure that enables knowledge transfer across systems and the distinctions in the underlying processes governing each individual system.


Retrieval-augmented approaches offer a natural path toward this goal by identifying relevant systems and leveraging their data to augment local predictions with information drawn from the broader system population. 
These approaches have proven effective in language modeling~\cite{hu2024rag, gupta2024comprehensive, peng2024graphrag} and, more recently, in general time-series forecasting~\cite{han2025retrieval, liu2024retrieval} and environmental modeling~\cite{luo2025learning, yu2025rag}. 
The typical approach is to 
create data embeddings that summarize each scenario's characteristics and then retrieve other scenarios with similar embeddings. 
However, such conventional embedding-based retrieval can be limited by the quality of embeddings learned from sparse data. Additionally, 
embedding similarity does not guarantee consistency of underlying physical processes, since scenarios with a similar response may arise from different mechanisms. For example, a shallow lake may warm rapidly because of its low thermal mass, while a deeper lake in a warmer climate may reach similar temperatures through a different radiative balance. Although their embeddings may be similar, the underlying flux dynamics are not. Retrieving such a scenario can introduce dynamics that conflict with the target system and degrade prediction performance.
This raises a question: How can we leverage physical knowledge to enhance embedding-based retrieval, ensuring that retrieved samples remain physically consistent with the target scenario's dynamics? 

To address this challenge, we propose \textbf{P}hysics-\textbf{I}nformed \textbf{E}nvironmental \textbf{R}etrieval (\textbf{PIER}), a retrieval framework for environmental time-series modeling.
In addition to the standard embedding-based retrieval, PIER introduces a physics-aware retrieval stream that scores candidates by 
how similarly they respond to physical fluxes relative to the target, using local verifiers trained on physics-derived flux features. 
The two retrieval streams are complementary: embedding similarity uses embeddings extracted by a global model and  captures shared temporal patterns of environmental drivers, while local physical consistency identifies candidates whose underlying dynamics match those of the target. 
However, their relative reliability varies across scenarios, e.g., systems with sparse local observations may depend more on embedding similarity, whereas well-observed systems can build reliable local verifiers and place greater weight on physical consistency. 
PIER addresses this through a weight adjustment mechanism that learns per-scenario weights to adaptively balance the two streams.

We evaluate the proposed method for modeling two important water quality variables, water temperature and dissolved oxygen (DO) dynamics using a comprehensive dataset of 356~lakes across the Midwestern United States spanning 41~years.
These lakes differ in size, depth, morphometry, and ecological state, while observations vary dramatically in coverage. These two variables are key indicators of water quality and ecosystem health~\cite{wilson2010water,solomon2013ecosystem,phillips2020time}. Water temperature regulates metabolic rates, chemical reactions, and oxygen solubility, shaping overall ecosystem dynamics~\cite{staehr2010lake}. DO concentrations are crucial for aquatic biodiversity and water safety since oxygen fluctuations in lakes often reflect their ecological balance and overall health~\cite{hanson2003lake, birge1906gases}. 
Our contributions are:
\begin{itemize}
	\item We propose PIER, a model-agnostic retrieval framework that augments embedding-based retrieval with a physics-aware stream using local verifiers, ensuring retrieved scenarios are both broadly relevant and physically grounded.
	\item We develop a weight adjustment mechanism that adaptively balances the two retrieval streams per scenario based on the reliability of local verifiers.
	\item We validate PIER across 356 real-world freshwater lakes, demonstrating consistent improvements over strong baselines.
\end{itemize}

\section{Problem Definition and Background}
\label{sec:prelim}

\subsection{Problem Definition.}
We consider a collection of environmental systems indexed by $\ell\in \{1,\dots,L\}$, each observed over a particular time period. 
For each target system, we have access to its environmental features $\mathbf{x}_t$ at each time $t$, which may include geographic and morphometric attributes, meteorological forcing, ecological indicators, and land-use characteristics. 
In addition to these inputs, the prediction target~$y_t$ is observed on a subset of days, where the number and temporal distribution of observations vary substantially across systems.

In general, a scenario is a fixed-length environmental context drawn from a single system, capturing its local conditions and observations. 
In our lake dataset, we define a scenario as a yearly data segment from a specific lake.
Multiple scenarios may originate from the same system (e.g., different years of the same lake) or different systems. Retrieval is performed over all scenarios in the training set.
Our objective is to predict the target variable at all time steps for scenarios in the testing period. Because observations 
are typically insufficient to capture the dynamics of each system, 
we aim to retrieve the most
relevant scenarios from all available data and use them to adapt a scenario-specific predictor. The core challenge is defining relevance in a way that accounts for both temporal similarity and physical consistency with the target system's dynamics.


\begin{figure*}[t]
  \centering
  \centerline{\includegraphics[width=0.95\linewidth]{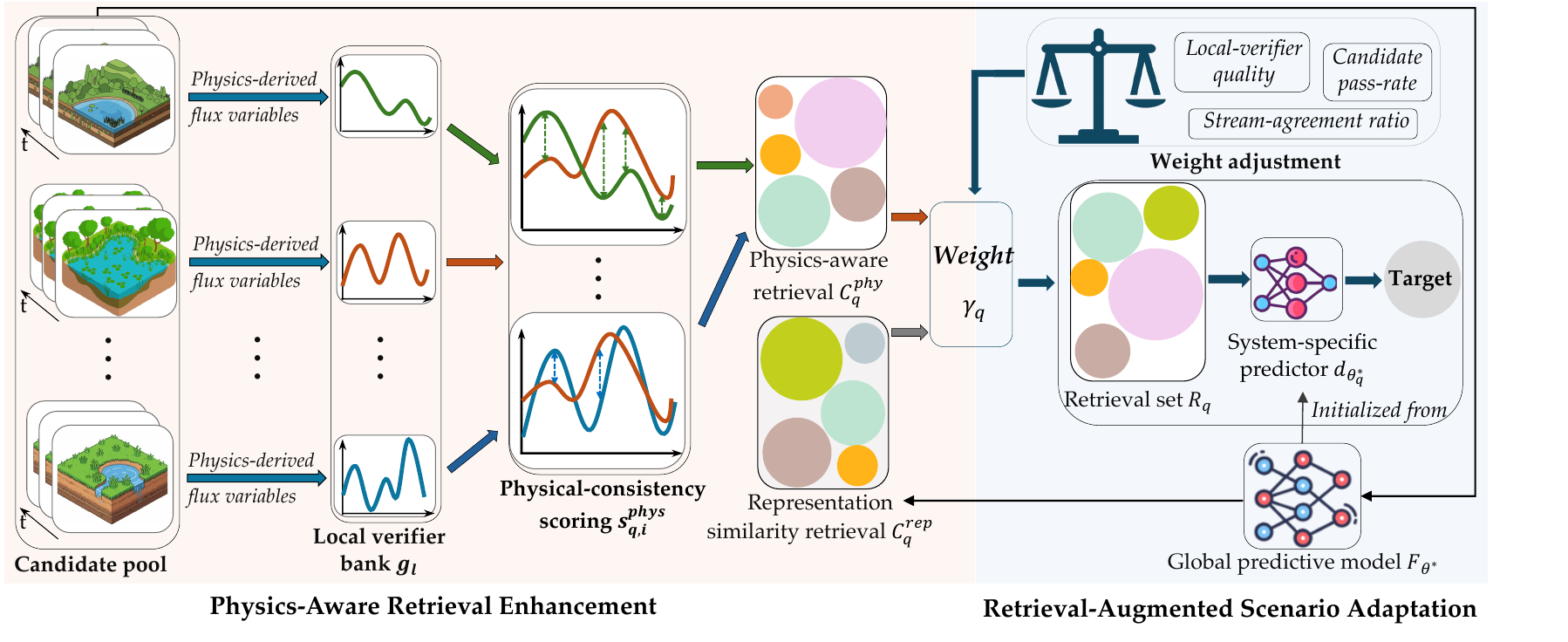}}
  \vspace{-0.3cm}
  \caption{Overview of PIER.}
  \label{fig:pipeline}
\end{figure*}

\subsection{Physics-Derived Data.}
In many environmental domains, process-based models are available that simulate system behavior based on established physical laws. 
Although limited by simplified parameterizations~\cite{jia2019physics},
these models produce additional physics data that PIER can leverage: (1) globally calibrated simulated labels that compensate for sparse observations, and (2) intermediate physics-derived flux features for each system, which serve as inputs to local dynamics models described in~\S\ref{sec: local bank}. These flux features are distinct from the environmental input features~$\mathbf{x}_t$: while input features describe external conditions (e.g., meteorological forcing), flux features capture the system’s internal physical response to those conditions.

\section{Method.}
\label{sec:method}

In the following, we first provide an overview of the PIER framework, and then describe the physics-aware retrieval enhancement process and the retrieval-augmented adaptation process. 

\subsection{Framework Overview.}
\label{sec:overview}
The PIER framework is built on two complementary ideas: (1) environmental systems of the same type share universal underlying patterns of input-response relationships, 
and (2) each system also exhibits unique dynamics, as certain variables in the universal relationships are not observable and thus cannot be included as input features.  
PIER consists of a
global predictive model and a retrieval-augmented adaptation mechanism. The global predictive model $\mathcal{F}_{\theta^*}$ is trained across all available scenarios using environmental input features, while mapping each scenario to a latent representation.  
The retrieval could be performed based on the representation similarity,  
which captures the shared response patterns learned from available data. 
However, relying on data embeddings alone does not guarantee that the retrieved systems share similar underlying physical processes. 

As shown in Fig.~\ref{fig:pipeline}, PIER addresses this through two key mechanisms:
\textbf{(1)~Physics-aware retrieval enhancement:} local verifiers  trained on physics-derived features score candidates by whether they share  flux-response dynamics consistent with the target system. The \textit{representation similarity stream} and \textit{physics-aware  stream} produce a merged candidate pool that is both broadly relevant and physically grounded.
\textbf{(2)~Retrieval-augmented scenario adaptation}: because the physics-aware stream's reliability varies across scenarios, a learned gate adaptively balances the two streams for the target scenario, and the top-ranked scenarios are used to adapt a scenario-specific predictor.

\subsection{Physics-Aware Retrieval Enhancement.}
\label{sec:physics_retrieval}

Rather than relying on a single retrieval criterion, we draw candidates 
from both streams and merge them.
We use $q$ to denote the \emph{query scenario}, and $i$ to index \emph{candidate scenarios}. 
$\text{Top-}K(\cdot)$ means selecting the $K$ candidates with the highest scores, where $K$ is a hyperparameter controlling the size of the retrieval set.
For each query $q$, this process can be expressed as
\begin{equation}
\small
\begin{gathered}
\mathcal{C}^{\text{rep}}_q = \text{Top-}K\big(s^{\text{rep}}_{q,i}\big),
\qquad
\mathcal{C}^{\text{phys}}_q = \text{Top-}K\big(s^{\text{phys}}_{q,i}\big), \\
\mathcal{U}_q = \mathcal{C}^{\text{rep}}_q \cup \mathcal{C}^{\text{phys}}_q ,
\end{gathered}
\end{equation}
where $\mathcal{C}^{\text{rep}}_q$ contains the top candidates ranked by representation similarity $s^{\text{rep}}_{q,i} = \cos(e_q, e_i)$, with $e_q$ being the global model's hidden-state representation for the scenario~$q$; and $\mathcal{C}^{\text{phys}}_q$ contains the top candidates ranked by a physical-consistency score $s^{\text{phys}}_{q,i}$ defined below.
The representation-similarity stream provides broad, stable candidates, preferring those that share general patterns but may lack physical relevance; the physics stream identifies candidates that are locally grounded in the target's underlying dynamics.
We now describe how the physical-consistency score $s^{\text{phys}}_{q,i}$ is constructed.

\subsubsection{Local dynamics verifier bank.}
\label{sec: local bank}
For each system~$\ell \in \{1, \dots, L\}$, we create a 
local verifier~$g_\ell$ using only the physics-derived flux variables from that system. Unlike the global model, which is trained using the full observation data across all systems, each local dynamics verifier is trained independently for each system. The input to this model includes only those flux variables that are relevant to the final target variable. These flux variables are generated by an independent process-based model.
While relying on such a small set of selected flux variables may reduce the ability of model~$g_\ell$ to fit observation data as closely as the pure data-driven model, it can better capture system-specific underlying patterns and may be less susceptible to overfitting.

\subsubsection{Physical-consistency scoring.}
Ideally, if a candidate scenario shares similar physical dynamics with the target scenario, then the target system's local verifier 
should be able to predict the candidate's behavior well. Conversely, if the candidate is governed by different relationships, the target's model will produce poor predictions. 
Given a target scenario from system~$\ell_q$, we score each candidate scenario~$i$ by applying the target system's local verifier to the candidate's flux features and measuring how well the prediction aligns with the candidate's observed targets:
\begin{equation}
\small
s^{\text{phys}}_{q,i} = \exp\left(
-\frac{1}{\tau}
\sqrt{
\frac{1}{|\mathcal{T}_i|}
\sum_{t \in \mathcal{T}_i}
\left(
g_{\ell_q}(\mathbf{x}^{\text{flux}}_i)_t - y^{(i)}_t
\right)^2
}
\right),
\label{eq:lake_score}
\end{equation}
where $\mathbf{x}_i^{\text{flux}}$ denotes the physics-derived flux features for candidate scenario $i$ and $\tau$ is a scaling parameter, and $\mathcal{T}_i$ denotes the set of time steps for which ground-truth observations are available for candidate scenario $i$.
The exponential mapping converts the prediction error into normalized similarity range $s^{\text{phys}}_{q,i}\in(0,1]$, ensuring that small errors yield scores close to 1 and large errors decay smoothly toward 0, which is preferable to a hard threshold.
A candidate whose dynamics are well reproduced by the target’s local verifier receives a high score, whereas poor predictability indicates physical inconsistency.

\subsubsection{Model-quality safeguard.}
\label{sec:safeguard}
While the local verifiers ${g_\ell}$ are intended to capture system-specific patterns, some may be less reliable due to limited
  observations for certain systems and biases in process-based model estimates of input flux variables. 
We assess each model's quality by defining $\mu_\ell = \mathcal{L}_{\text{sim},\ell} - \mathcal{L}_{\text{model},\ell}$, where $\mathcal{L}_{\text{sim},\ell}$ and $\mathcal{L}_{\text{model},\ell}$ are the prediction errors of the process-based simulator and the learned model on system~$\ell$'s held-out validation data, respectively.
When $\mu_\ell > 0$, the learned model outperforms the simulator, indicating reliable local dynamics have been captured. Otherwise, we disable the physical stream for that target scenario and rely on representation similarity alone.

\subsection{Retrieval-Augmented Scenario Adaptation.}
\label{sec:adaptation}

The physics-aware  stream enhances retrieval by introducing candidates that are grounded in the target's local dynamics. However, each environmental system is unique, and this stream is not equally reliable across all scenarios. For a system with abundant observations and a well-fitted local verifier, physical-consistency scores are reliable signals and should carry substantial weight in reranking. For a system with a poorly calibrated local verifier, those scores become noisy, and relying on them can degrade retrieval quality. Uniformly weighting candidates in the merged pool cannot accommodate this variability.
PIER addresses this by adaptively determining 
how much the physics-aware  stream contributes to the target scenario through a dynamic scoring mechanism. Then it uses the reranked retrieval set to create a scenario-specific predictor.


\subsubsection{Combined retrieval score.}
For each candidate~$i$ in~$\mathcal{U}_q$, we combine the representation similarity~$s^{\text{rep}}_{q,i}$ with the physical-consistency~$s^{\text{phys}}_{q,i}$:
\begin{equation}
\small
s^{\text{comb}}_{q,i} = (1 - \gamma_q)\, s^{\text{rep}}_{q,i} + \gamma_q\, s^{\text{phys}}_{q,i}, \quad \gamma_q \in [0,1].
\label{eq:rerank}
\end{equation}
where $\gamma_q$ controls how much the physics-aware stream influences the final ranking for this scenario.
When $\gamma_q = 0$, retrieval relies entirely on representation similarity; as $\gamma_q$ increases, physical consistency plays a larger role. 
The top-$K$ candidates ranked by $s^{\text{comb}}_{q,i}$ form the retrieval set $\mathcal{R}_q$. The weight parameter $\gamma_q$ can be automatically adjusted using the diagnostic features as described later. 

\subsubsection{Scenario-adaptive prediction.}
The final predictions for each testing scenario are generated by a system-specific predictor~$d_{\theta_q^\star}$ initialized from the global predictive model and fine-tuned using the retrieved data. 
Starting from the global model (parameterized by $\theta^\star$ trained over all samples ) provides a strong starting point that already captures general environmental dynamics. 
Fine-tuning is performed using sparse observations from retrieved scenarios, as follows: 
\begin{equation}
\small
\mathcal{L}_{\text{adapt}} =
\sum_{i \in \mathcal{R}_q} w_i
\frac{1}{|\mathcal{T}_i|}
\sum_{t \in \mathcal{T}_i}
\left(
\hat{y}^{(i)}_t - y^{(i)}_t
\right)^2,
\label{eq:adapt}
\end{equation}
where $\mathcal{T}_i$ denotes the set of time steps with available observations for retrieved scenario $i$, the weight $w_i$ measures the contribution made by each retrieved scenario to the target scenario $q$. 
Given a retrieval set~$\mathcal{R}_q$ with associated weights, each retrieved scenario contributes in proportion to its normalized combined score:
\begin{equation}
\small
w_i = \frac{s^{\text{comb}}_{q,i}}{\sum_{j \in \mathcal{R}_q} s^{\text{comb}}_{q,j}}.
\end{equation}

The global predictive model $\mathcal{F}_{\theta^*}$ and system-specific predictor $d_{\theta_q^\star}$ are designed to be model-agnostic so that different network backbones can be plugged in seamlessly.  
In our main experiments, we use a long short-term memory (LSTM) network as the default backbone, given its strong performance on sparse hydrological and aquatic time series~\cite{hochreiter1997long}. We verify that the framework generalizes across architectures in~\S\ref{sec:enhanced}.

\subsubsection{Weight adjustment using diagnostic features.}
The quality of retrieval is ultimately measured by the quality of the adapted predictor.
This naturally gives rise to a bilevel optimization: the \emph{inner level} adapts the predictor~$d_{\theta_q^\star}$ on the retrieval set (Eq.~\ref{eq:adapt}), while the \emph{outer level} selects the degree of physics enhancement~$\gamma_q$ that yields the best adapted predictor.
A choice of~$\gamma_q$ determines the combined scores (Eq.~\ref{eq:rerank}), which determine the retrieval set~$\mathcal{R}_q$ and the weights~$\{w_i\}$, which determine the inner-level adaptation, and ultimately the prediction quality on the target scenario.
\begin{equation}
\small
\mathcal{L}_{\text{outer}} =
\frac{1}{|\mathcal{T}_q|}
\sum_{t \in \mathcal{T}_q}
\left(
d_{\theta_q^*(\gamma_q)}(\mathbf{x_q})_t - y^{(q)}_t
\right)^2,
\end{equation}
where $\theta_q^\star(\gamma_q)$ denotes the predictor parameters obtained after the inner-level adaptation on the retrieval set induced by~$\gamma_q$.

Directly optimizing~$\gamma_q$ by differentiating through the inner-level adaptation would require backpropagating through an entire training procedure, which requires unrolling the optimization or computing second-order gradients. We instead adopt a two-phase approach.
In the first phase, for each training scenario~$q$, we evaluate the full bilevel pipeline across a discrete set of~$\gamma$ values: for each candidate~$\gamma$, we run the complete inner-level adaptation and compute~$\mathcal{L}_{\text{outer}}$.
We record the value that achieves the lowest outer-level loss as the per-scenario target:
\begin{equation}
\small
  \gamma_q^\star
  = \arg\min_{\gamma \in \mathcal{G}}\;
    \mathcal{L}_{\text{outer}}(\gamma),
  \label{eq:best_gamma}
\end{equation}
where $\mathcal{G} \subset [0,1]$ is a discrete grid.
In the second phase, we train a small MLP gate~$h_\phi$ to predict these optimal weight values. This two-phase design sidesteps the cost of differentiating through the inner loop while still optimizing the gate for downstream prediction quality. At test time, the gate produces~$\hat{\gamma}_q$ in a single forward pass, making the full pipeline efficient at inference. 
Specifically, the weight parameter is predicted 
from diagnostic features $\mathbf{z}_q$:
\begin{equation}
\small
  \hat{\gamma}_q = h_\phi(\mathbf{z}_q),
  \qquad
  \mathcal{L}_{\text{gate}}
  = \big(\hat{\gamma}_q - \gamma_q^\star\big)^2,
  \label{eq:gate}
\end{equation}
where the diagnostic feature vector~$\mathbf{z}_q$ captures signals about each stream's reliability for the current scenario: 
\textit{Local-verifier quality}: the model-quality margin $\mu_{\ell_q}$ from~\S\ref{sec:safeguard}, measuring how well the target's local verifier performs on its own system.
\textit{Candidate pass-rate}: the fraction of candidates in~$\mathcal{U}_q$ for which the target's local verifier~$g_{\ell_q}$, applied to the candidate's flux features, predicts the candidate observations with lower error than the simulation, measuring whether the local verifier useful for scoring the current candidates. 
\textit{Stream-agreement ratio}: the fraction of overlap between  $\mathcal{C}^{\text{rep}}_q$ and  $\mathcal{C}^{\text{phys}}_q$. High overlap suggests both streams agree and the physics enhancement is reinforcing the representation-similarity ranking.

\begin{figure}
	\centering
	\centerline{\includegraphics[width=0.8\linewidth]{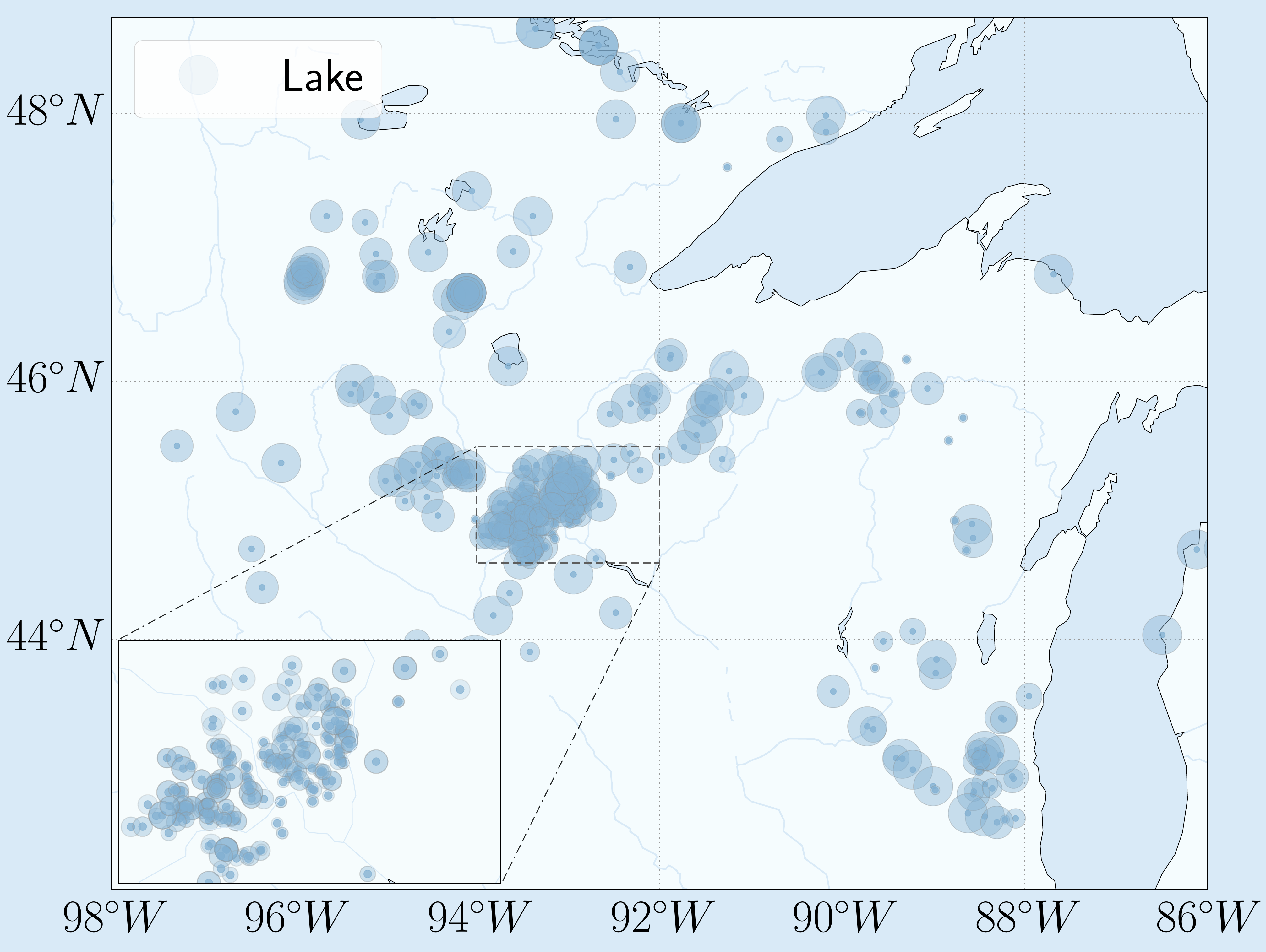}}
    \vspace{-0.2cm}
	\caption{Map of 356 tested lakes. } 
	\label{fig:map}
\end{figure}

\begin{table*}[t]
	\centering
    \small
	\caption{Comparative RMSE performance.  Lower values are better; grey numbers denote standard deviation.}
	\label{tab:rmse_results}
    \begin{tabular}{p{2cm}cccccc}
		\toprule
		\multirow{2}{*}{Algorithm} &
		\multicolumn{3}{c}{\textbf{DO concentration ($ \bf{g / m^{3}}$)}} &
		\multicolumn{3}{c}{\textbf{Water temperature ($^\circ$C)}} \\
		\cmidrule(lr){2-4} \cmidrule(lr){5-7}
		& Summer (epi.) & Summer (hyp.) & Fall to spring & Summer (epi.) & Summer (hyp.) & Fall to spring \\
		\midrule
		Physics & 2.1018\,\textcolor{gray}{(0.000)} & 2.8793\,\textcolor{gray}{(0.000)} & 2.7438\,\textcolor{gray}{(0.000)} & 1.6349\,\textcolor{gray}{(0.000)} & 2.7821\,\textcolor{gray}{(0.000)} & 1.5436\,\textcolor{gray}{(0.000)}  \\
		LSTM          & 1.9841\,\textcolor{gray}{(0.011)} & 2.9404\,\textcolor{gray}{(0.106)} & 2.2172\,\textcolor{gray}{(0.024)} & 1.5043\,\textcolor{gray}{(0.036)} & 1.8524\,\textcolor{gray}{(0.117)} & 1.2679\,\textcolor{gray}{(0.067)}  \\
		Informer       & 2.0903\,\textcolor{gray}{(0.016)} & 3.2290\,\textcolor{gray}{(0.034)} & 2.0687\,\textcolor{gray}{(0.040)} & 1.9545\,\textcolor{gray}{(0.033)} & 3.1358\,\textcolor{gray}{(0.305)} & 2.0768\,\textcolor{gray}{(0.163)} \\
		iTransformer  & 2.5965\,\textcolor{gray}{(0.416)} & 3.4620\,\textcolor{gray}{(0.060)} & 2.5482\,\textcolor{gray}{(0.288)} & 2.2468\,\textcolor{gray}{(0.179)} & 2.4872\,\textcolor{gray}{(0.094)} & 2.0218\,\textcolor{gray}{(0.109)}  \\
		TSMixer       & 2.2486\,\textcolor{gray}{(0.042)} & 3.0824\,\textcolor{gray}{(0.014)} & 2.3996\,\textcolor{gray}{(0.062)} & 2.2038\,\textcolor{gray}{(0.034)}& 1.9521\,\textcolor{gray}{(0.064)} & 1.9115\,\textcolor{gray}{(0.067)}  \\
		TimeMixer     & 2.7571\,\textcolor{gray}{(0.101)} & 3.4909\,\textcolor{gray}{(0.035)} & 2.6167\,\textcolor{gray}{(0.107)} & 2.6866\,\textcolor{gray}{(0.131)}& 3.2544\,\textcolor{gray}{(0.162)} & 2.5377\,\textcolor{gray}{(0.137)}  \\
		TimesNet      & 2.0879\,\textcolor{gray}{(0.064)} & 2.7629\,\textcolor{gray}{(0.029)} & 2.1690\,\textcolor{gray}{(0.015)} & 2.0766\,\textcolor{gray}{(0.059)}& 1.9501\,\textcolor{gray}{(0.060)} & 2.0817\,\textcolor{gray}{(0.023)}  \\
		\textbf{PIER} & \textbf{1.8888}\,\textcolor{gray}{(0.004)} & \textbf{2.5395}\,\textcolor{gray}{(0.002)} & \textbf{2.0605}\,\textcolor{gray}{(0.040)} & \textbf{1.2486}\,\textcolor{gray}{(0.010)} & \textbf{1.7020}\,\textcolor{gray}{(0.012)} & \textbf{1.0673}\,\textcolor{gray}{(0.039)}  \\
		\bottomrule
	\end{tabular}
\end{table*}

\begin{figure*}[t]
	\centering
	\centerline{\includegraphics[width=\linewidth]{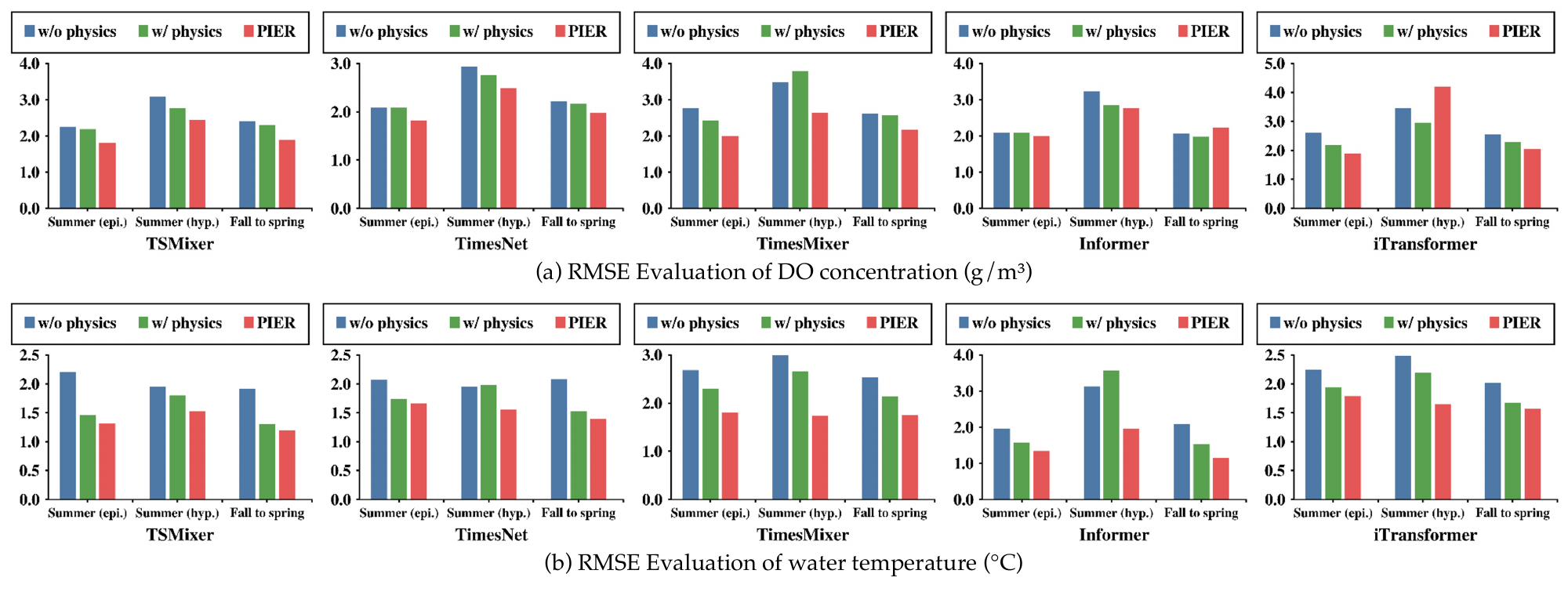}}
    \vspace{-0.4cm}
	\caption{Comparison under three training settings. } 
    \vspace{-0.3cm}
	\label{fig:enhanced}
\end{figure*}

\section{Experimental Evaluation.}
We conduct extensive experiments across various lakes in the Midwestern United States to address research questions:

\begin{itemize}
  \item \textbf{RQ1.} How does PIER compare to other temporal prediction methods in predictive performance?
  \item \textbf{RQ2.} Can PIER serve as a general augmentation strategy that improves diverse model backbones?
  \item \textbf{RQ3.} 
  How does physics-aware retrieval enhance capturing temporal patterns in water temperature and DO concentration? 
  \item \textbf{RQ4.} What is the contribution of each component in PIER's pipeline?
\end{itemize}

\subsection{Data Preparation.}
We evaluate PIER's predictive performance for water temperature and DO concentrations using a comprehensive dataset spanning 41 years (1979–2019), with data up to 2011 for training, 2012–2015 for validation, and 2016–2019 for testing. The dataset comprises ecological observations from 356 lakes across the Midwestern USA. The spatial distribution of these lakes is shown in Fig.~\ref{fig:map}, where color intensity represents lake depth, and marker size corresponds to surface area. 
The dataset contains approximately 1.75 million daily records, each including 47 environmental features such as morphometric attributes, meteorological conditions, trophic states, and land use characteristics. Data sources are described in~\cite{meyer2024national,yu2024adaptive,willard2021predicting}. Specifically, water temperature data were obtained from the U.S. Geological Survey~\cite{willard2021predicting}, with 476,215 observed measurements recorded across depths on 57,156 distinct days. DO concentration data were retrieved from the Water Quality Portal (WQP), encompassing 23,192 days of observations. Lake residence time was sourced from the HydroLAKES dataset. Trophic state probabilities were derived from~\cite{meyer2024national}. Land use proportions within each lake's watershed were extracted from the National Land Cover Database (NLCD). 

\paragraph{Domain context:}
Lakes undergo seasonal stratification driven by thermal expansion of water. During summer, a stable vertical density gradient separates the water column into the epilimnion (a warm, well-mixed surface layer) and the hypolimnion (a cooler, denser bottom layer)~\cite{read2011derivation}. This stratification restricts vertical mixing, limiting oxygen exchange between layers.
From fall through spring, the water column becomes well-mixed, behaving as a single layer.
This seasonal structure motivates our prediction targets: during summer, we predict temperature and DO separately for the epilimnion (\textbf{epi.}) and hypolimnion (\textbf{hyp.}); from fall to spring, when the water column is mixed, we predict the total DO concentration throughout the lake.

\paragraph{Physics-derived data:}
For water temperature, we employ the General Lake Model~\cite{hipsey2019general}, which simulates the principal heat energy fluxes governing lake thermal dynamics, including incoming short-wave radiation, long-wave radiation, back radiation, sensible heat flux, and latent evaporative heat flux. For DO concentration, we use the process-based model~\cite{ladwig2022long}, which simulates atmospheric exchange, net ecosystem production, sediment oxygen demand, and turbulent entrainment fluxes between layers during summer stratification. These models produce 
globally calibrated simulation labels for pretraining the global predictive
model and physics-derived flux features for training the local verifiers.

\subsection{Baselines.} 
To evaluate the effectiveness of our PIER framework, we compare it against several baseline models commonly used in time series forecasting: \textbf{Physics}~\cite{hipsey2019general,ladwig2022long}: It represent state-of-the-art physics-based models.
\textbf{LSTM}~\cite{hochreiter1997long}: A recurrent neural network architecture designed to capture long-term dependencies in sequential data, widely used for time series forecasting.
\textbf{Informer}~\cite{zhou2021informer}: A forecasting model for very long time-series data that keeps the Transformer’s ability to learn long-range patterns while making it fast and memory-efficient enough to predict long future sequences in one shot.
\textbf{iTransformer}~\cite{liuitransformer}: A Transformer-based model designed for irregular time series data, incorporating adaptive attention mechanisms to better capture temporal dependencies.
\textbf{TSMixer}~\cite{chen2023tsmixer}: An all-MLP architecture for time series forecasting that integrates historical data, future known inputs, and static contextual information through conditional feature mixing and mixer layers.
\textbf{TimesNet}~\cite{wutimesnet}: A framework that transforms one-dimensional time series into two-dimensional representations, allowing the use of 2D convolutional kernels to extract temporal patterns.
\textbf{TimeMixer}~\cite{wangtimemixer}: A model that explores mixing-based architectures for time series forecasting, focusing on effectively capturing short- and long-term dependencies.

\subsection{Implementation Details.} 
Hyperparameters for all methods were tuned via grid search, with optimal configurations selected on validation performance.
We explored batch sizes in $\{8, 16, 32\}$, learning rates in $[0.001, 0.05]$, and hidden dimensions in $[20, 200]$.
For Transformer-based models, the number of attention heads was varied in $\{4, 8, 16\}$.
For PIER, we additionally tuned the retrieval set size~$K$, the scaling parameter~$\ tau$ in Eq.~\ref{eq:lake_score}. We set $K = 50$ and $\tau = 10$.

\subsection{Experimental Results} 
\subsubsection{Performance comparison.}
Table~\ref{tab:rmse_results} reports the RMSE performance across all baselines for both DO concentration and water temperature prediction under different seasonal regimes. PIER achieves the best performance across all six settings, consistently outperforming both physics-based and data-driven approaches:
\begin{enumerate}[label=(\alph*)]
\item Increasing model complexity does not help.
Advanced architectures designed for complex temporal patterns, such as iTransformer, TSMixer, etc, all underperform the
simpler LSTM on this dataset. These models are designed for data-rich settings and appear to struggle with the severe sparsity characteristic of environmental observations. 
\item PIER consistently improves over all baselines across all six settings, with the largest gains in the most challenging cases such as hypolimnetic DO in summer. This suggests the bottleneck lies in how data are used rather than in the model's complexity. PIER addresses this through a data-centric view, retrieving and reusing informative scenarios filtered through physical-consistency constraints.
\end{enumerate}

\begin{figure}[t]
	\centering
	\centerline{\includegraphics[width=0.95\linewidth]{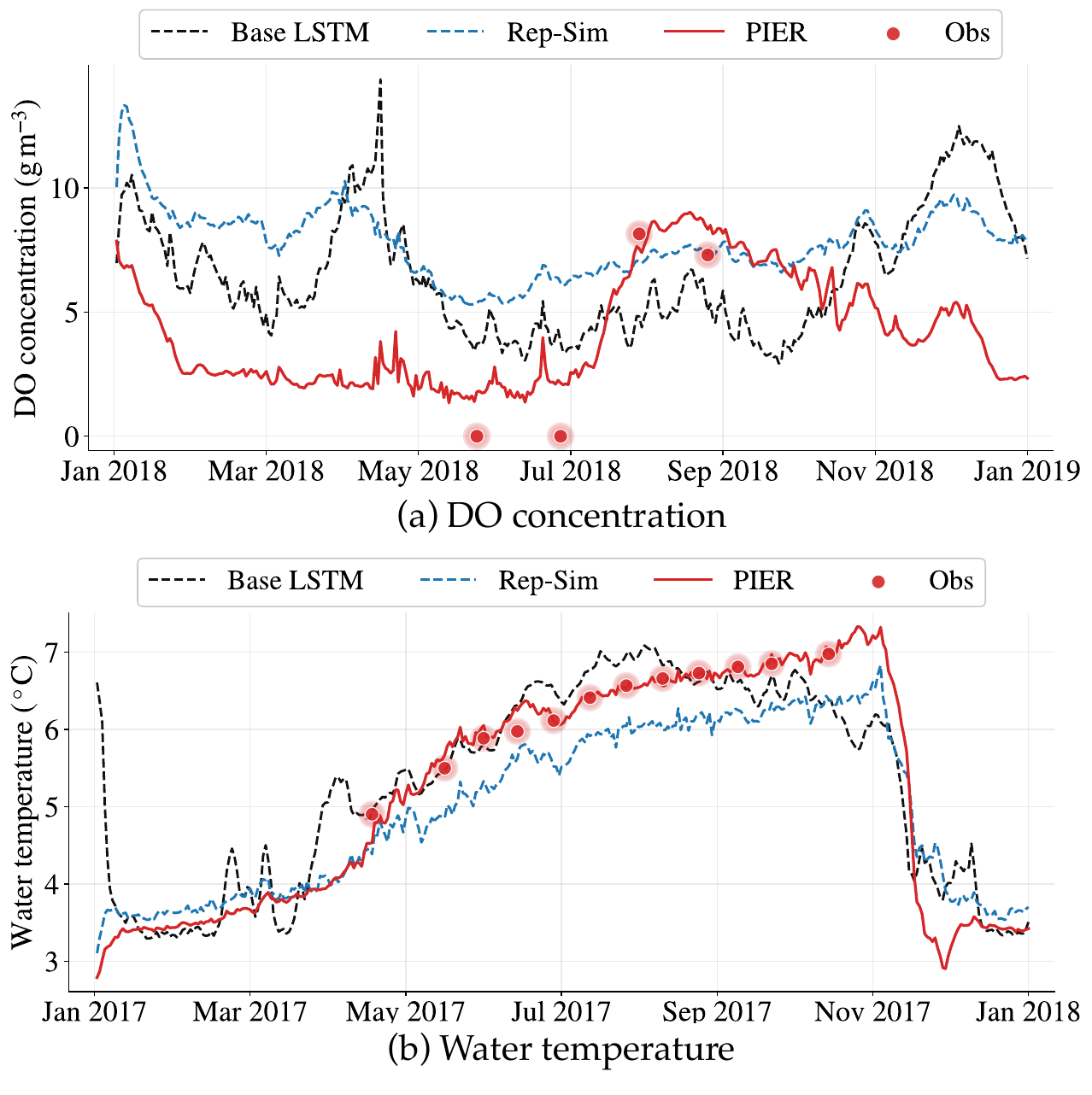}}
    \vspace{-0.3cm}
	\caption{Time-series predictions for two case study scenarios. (a) Epilimnion DO concentration for \textit{Long Lake}, Year 2018. (b) Hypolimnion water temperature for \textit{Weaver Lake}, Year 2017.} 
	\label{fig:case_timeseries}
\end{figure}

\begin{figure}[t]
	\centering
	\centerline{\includegraphics[width=0.95\linewidth]{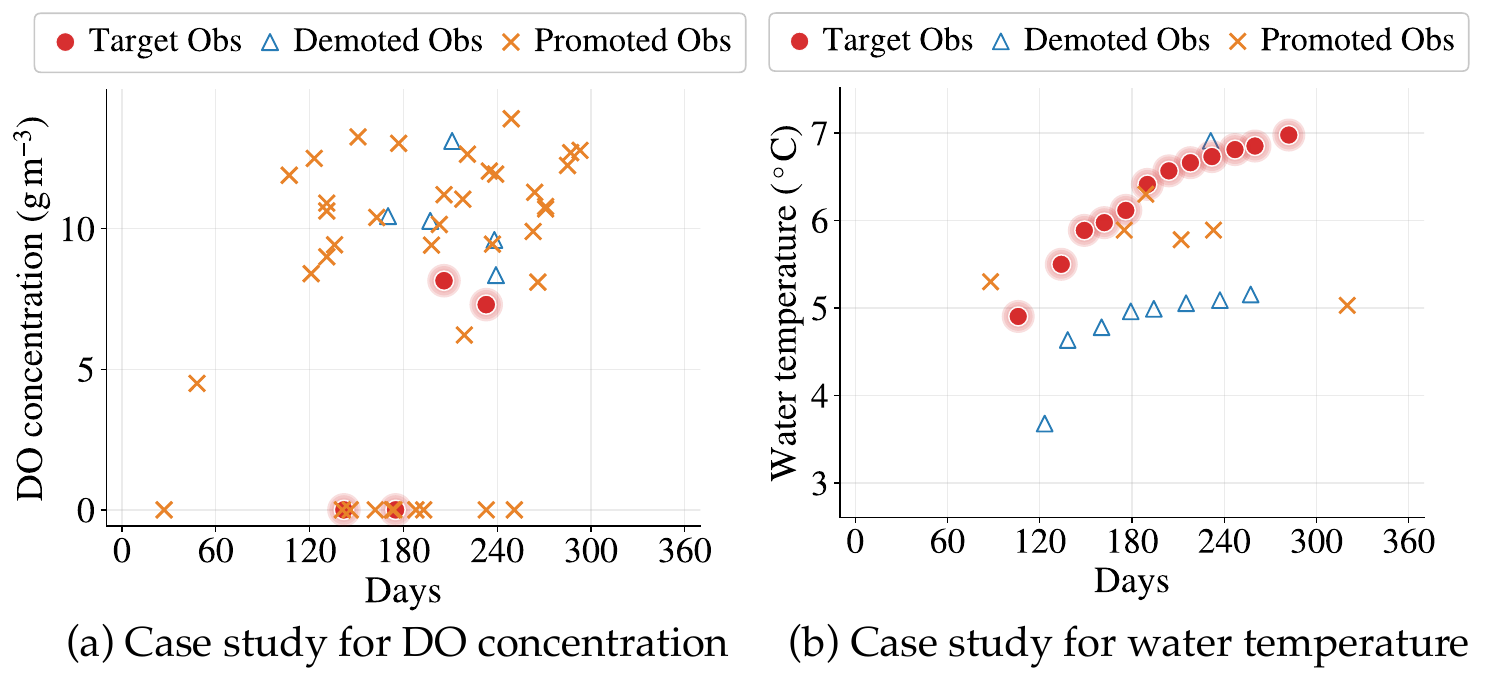}}
    \vspace{-0.3cm}
	\caption{Observations of promoted and demoted candidates alongside target observations. Promoted candidates are shown in orange; demoted candidates in blue. Observations are plotted against day of year.} 
	\label{fig:case_obs}
\end{figure}

\subsubsection{PIER as a general augmentation strategy.}
\label{sec:enhanced}

To assess the broader applicability of PIER, we further integrate it with five predictive models and compare each model's performance under three settings (RQ2):
(1)~trained only on observed labels with standard environmental features,
(2)~pretrained on the simulation labels produced by the process-based model and fine-tuned on observations, with physics-derived flux features concatenated to the standard inputs, and
(3)~the full PIER framework.
Fig.~\ref{fig:enhanced} reports the results.
\begin{enumerate}[label=(\alph*)]
\item Physics enhancement outperforms direct  training across the majority of tested models and tasks, confirming that the simulation labels and flux features are useful resources for addressing data sparsity challenge in environmental systems. In a few cases, physics enhancement does not help, as the physics flux features, designed to capture system-specific physical responses, may not be fully exploitable when simply concatenated as additional inputs. This suggests PIER's local verifiers provide a more principled way to leverage them.
\item PIER achieves the lowest RMSE across almost all tasks, demonstrating its applicability across diverse model architectures. While physics-derived data improves models by leveraging better data sources, the retrieval mechanism provides additional gains. In tasks such as DO prediction with iTransformer, PIER does not improve over the physics-enhanced setting, as the DO hyp. dynamics are particularly challenging for this backbone, and fine-tuning on retrieved scenarios can amplify rather than correct the base model's weaknesses.
\end{enumerate}

\subsection{Case Study.}
To understand how PIER's retrieval mechanism works in practice and where the improvements come from, we examine two representative target scenarios in detail: one for DO concentration and one for water temperature.
For each case, we present time-series predictions by Base LSTM, Representation-Similarity Retrieval, and PIER in Fig.~\ref{fig:case_timeseries}, and observations of promoted candidates and demoted candidates alongside target observations in Fig.~\ref{fig:case_obs}.
Demoted candidates are those ranked high by representation similarity but down-weighted by PIER; promoted candidates are those elevated by the physics-aware stream.

\subsubsection{Case 1.}
Fig.~\ref{fig:case_timeseries}(a) and Fig.~\ref{fig:case_obs}(a) show epilimnion DO for \textit{Long Lake} in 2018, where the observed DO drops to near zero around July. Base LSTM and representation-similarity retrieval both predict 7–14 g/m³ throughout summer (RMSE 3.33 and 4.14, respectively); retrieval actually worsens the base model because the demoted candidates are dominated by the target lake's own historical years, which reflect typical high-DO behavior. Embeddings encode statistically typical patterns from training data and fail to generalize when test conditions deviate from what the model has seen. The physics-aware stream promotes candidates from other lakes (\textit{Bashaw Lake}-2009/2015, \textit{Parley Lake}-2006) with compatible flux-response dynamics and similarly low DO, reducing error to 1.52.

\subsubsection{Case 2.}
Fig.~\ref{fig:case_timeseries}(b) and Fig.~\ref{fig:case_obs}(b) show hypolimnion water temperature for \textit{Weaver Lake} in 2017. 
Here, representation-similarity retrieval selects lakes with shared general temporal structure but mismatched thermal regimes, since the representations capture shared statistical structure from training but not physical compatibility. PIER removes these and promotes scenarios such as \textit{Bavaria Lake-2013} and \textit{Upper Nemahbin Lake-2008}, both within the target's narrow range. Notably, none of these appeared in the representation-similarity retrieval set. 

\subsection{Ablation Study.}
Ablation experiments are designed to evaluate the contribution of each component while holding everything else
fixed: 
\textit{E0: Base LSTM}. No simulation and no physics flux features, no retrieval. The model is trained solely on observed labels.
\textit{E1: Simple retrieval}. Retrieval without simulated labels or physics flux features.
\textit{E2: Physics-aware stream only}. Full physics-aware retrieval enhancement pipeline, but ranking uses only the physical-consistency scores ($\gamma = 1$).
\textit{E3: Fixed equal weight}. Physics-aware stream and representation-similarity stream contribute equally with a fixed global weight ($\gamma = 0.5$). This tests whether a naive combination of the two streams already improves over either stream alone.
\textit{E4: Representation-similarity stream only}. The simulated labels and physics flux features are used, but ranking uses only the representation similarity scores ($\gamma = 0$).
\textit{E5: PIER}. Fixed $\gamma$ is replaced with the learned weight adjustment. 
Fig.~\ref{fig:ablation} presents the results:

\begin{figure}
	\centering
	\centerline{\includegraphics[width=\linewidth]{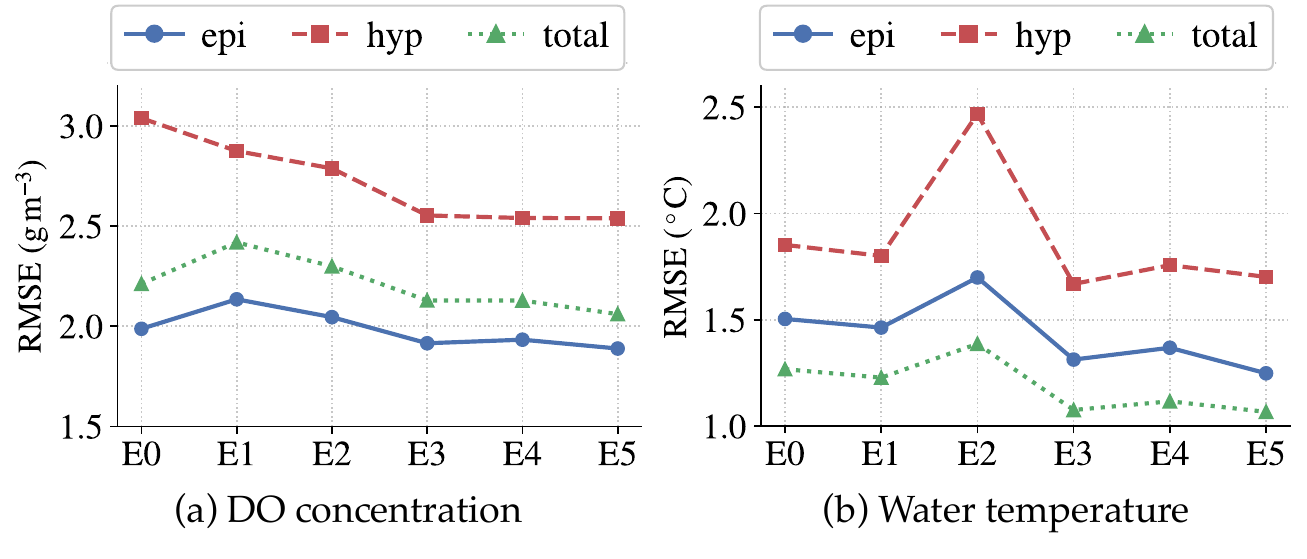}}
    \vspace{-0.3cm}
	\caption{Results of ablation study.
    } 
	\label{fig:ablation}
\end{figure}

\begin{enumerate}[label=(\alph*)]
\item Simple retrieval (E1) yields mixed results: it slightly improves water temperature predictions but actually degrades DO predictions. Without any physics components, retrieval can introduce misleading scenarios that hurt rather than help, particularly for complex DO dynamics. 

\item When the full physics components are used but ranking relies only on representation similarity (E4), performance improves substantially compared to E1. The improvement comes entirely from the physics-derived data that produces better representations to make retrieval more effective.

\item The Physics-aware retrieval alone (E2) improves over E1 for DO tasks but performs poorly on water temperature, as it can be noisy when used as the sole ranking criterion.
Notably, equal weighting (E3) improves over both E2 and E4, confirming the two streams provide complementary information.

\item PIER's learned weight adjustment (E5) achieves the best performance on all tasks, confirming that scenario-specific weight adjustment is necessary: no single fixed $\gamma$ is best across all settings, and neither is optimal for any individual scenario.
\end{enumerate}

\section{Conclusion.}
In this paper, we presented PIER, a physics-informed retrieval framework that augments embedding-based retrieval with a physics-aware stream grounded in local verifiers, and adaptively balances the two streams per scenario through a weight adjustment mechanism. Experiments on 356 lakes over 41 years show consistent gains over physics-based and data-driven baselines for water temperature and dissolved oxygen prediction, and demonstrate that PIER serves as a general augmentation strategy across diverse backbones. Future work includes extending PIER to other environmental domains, such as agriculture, and exploring additional ways to integrate physical processes into machine learning methods.


\newpage
\bibliographystyle{siamplain}
\bibliography{reference}

\end{document}